\documentclass[journal]{IEEEtran}
\usepackage[figuresright]{rotating}
\usepackage[linesnumbered,ruled,vlined]{algorithm2e}
\usepackage{algorithmic}
\usepackage{graphicx}
\usepackage{url}
\usepackage{amsmath}
\usepackage{subfigure}
\usepackage{threeparttable}
\usepackage{bm}
\usepackage{float}
\usepackage{color}
\usepackage{booktabs}  
\usepackage{multirow}
\usepackage{setspace}
\usepackage{caption}
\usepackage{epstopdf}
\usepackage{booktabs}
\usepackage{threeparttable}
\usepackage{pifont}
\usepackage{makecell}
\usepackage{cite}
\usepackage{amsfonts}

\begin{document}

\title{Deep Reinforcement Learning-based Multi-objective Path Planning on the Off-road Terrain Environment for Ground Vehicles}

\author{Shuqiao Huang,~\IEEEmembership{}
	Xiru Wu,~\IEEEmembership{}
	Guoming Huang ~\IEEEmembership{}% <-this % stops a space
	\thanks{Shuqiao Huang, Xiru Wu and Guoming Huang are with the School of Electronic Engineering and Automation, Guilin University of Electronic Technology, Guilin, Guangxi Province, China, e-mail: (\{xiruwu, huangguoming\}@guet.edu.cn).}% <-this % stops a space  yuanxiaofang@hnu.edu.cn
	
}

\markboth{Journal of \LaTeX\ Class Files,~Vol.~14, No.~8, August~2015}%
{Shell \MakeLowercase{\textit{et al.}}: Bare Demo of IEEEtran.cls for IEEE Journals}

\maketitle

\begin{abstract}

Due to the vastly different energy consumption between up-slope and down-slope, a path with the shortest length on a complex off-road terrain environment (2.5D map) is not always the path with the least energy consumption. For any energy-sensitive vehicle, realizing a good trade-off between distance and energy consumption in 2.5D path planning is significantly meaningful. In this paper, we propose a deep reinforcement learning-based 2.5D multi-objective path planning method (DMOP). The DMOP can efficiently find the desired path in three steps: (1) Transform the high-resolution 2.5D map into a small-size map. (2) Use a trained deep Q network (DQN) to find the desired path on the small-size map. (3) Build the planned path to the original high-resolution map using a path-enhanced method. In addition, the hybrid exploration strategy and reward shaping theory are applied to train the DQN. The reward function is constructed with the information of terrain, distance, and border. Simulation results show that the proposed method can finish the multi-objective 2.5D path planning task with significantly high efficiency. With similar planned paths, the speed of the proposed method is more than 100 times faster than that of the A* method and 30 times faster than that of H3DM method. Also, simulation proves that the method has powerful reasoning capability that enables it to perform arbitrary untrained planning tasks.

\end{abstract}
 
\begin{IEEEkeywords}
	2.5D Path planning, deep reinforcement learning, deep Q learning, reward shaping, ground vehicles
\end{IEEEkeywords}

% \fi
\IEEEpeerreviewmaketitle

\section{Introduction}

\subsection{Motivation}
\IEEEPARstart {T}{he} use of energy is a important topic of interest in the field of ground vehicles (GVs) \cite{2014Development}. GV such as exploration robot, space probe mostly run on the complex off-road terrain environment (2.5D environment), the energy consumption may sometimes become a problem that restricts their working range and time. So far, scholars have studied the battery technology \cite{affanni2005battery}, the motor control technology \cite{williamson2007comprehensive} or the gas engine technology \cite{stanton2013systematic}, in order to improve the energy efficiency.

In this paper, an energy-saving method from the perspective of path planning for GV is studied. A GV going uphill always consumes a lot of energy to overcome its gravity, which means a path with less up-slope but a bit longer, may save energy than that of the shortest path which covers lots of up-slopes. Hence, there will always be some energy-saving paths on the 2.5D map. Meanwhile, some of these paths with longer distances may significantly increase the traveling time, which is also unacceptable. Overall, that yields a multi-objective path planning problem: to find paths on terrain environment with good trade-off between energy consumption and distance. 

The traditional methods such as violent search \cite{Brandao7511785}, heuristic-based search \cite{Raja8023669} and probabilistic search \cite{Paton9341409}, etc. can solve this multi-objective path planning problem to some extend. However, all of them encounter some shortcomings on this task. The violent search methods have to search a large area with low efficiency. The heuristic-based methods suffer on the difficulty of modeling the heuristic function for 2.5D path planning (It is hard to estimate the cost of a path on the terrain map). The probabilistic search methods show unstable performance on the off-road environment. 

The DRL with the merit of solving complex programming problems has made a huge success in Go \cite{silver2017mastering}, video games \cite{mnih2015human}, Internet recommendation system \cite{munemasa2018deep}, etc. More and more institutes devote themselves to this area for the use of DRL. Among these, scholars also prove that DRL has the potential to solve the N-P hard problem \cite{fan2020finding}, such as traveling salesman problem (TSP) \cite{munemasa2018deep}, path planning on 2D map \cite{sichkar2019reinforcement}. Since the 2.5D path planning is also a kind of N-P hard problem, we develop a deep reinforcement learning-based multi-objective 2.5D path planning method (called DMOP) in this paper. %motivation

\subsection{Related Works}

Path planning methods that consider slope (on 2.5D map) have been studied for years, and the research focuses on finding suitable paths for GVs in complex terrain environments\cite{Usami8911549, Zhou8304280, Ding8816402, Paton9341409}. In the early research, the first thing to consider is the traversability of the path, so scholars consider the vehicle dynamics to confirm that the planned path will not cause the vehicle to get stuck, skid, roll over or be too steep to drive through the condition, focusing on whether it can be driven. For instances, in \cite{Paton9341409}, a probabilistic road map (PRM)-based path planning method considering the traversability is proposed for extreme-terrain rappelling rovers. \cite{Raja8023669} proposes a path planning method for a wheeled mobile robot operating in rough terrain dynamic environments using a combination of A* search algorithm and potential field method. \cite{Ugur9551617} investigates a reliable and robust rapidly exploring random tree (R2-RRT*) algorithm to tackle challenges in mission planning of off-road autonomous ground vehicles (AGVs) under uncertain terrain environment.

On the basis of ensuring traversability, many scholars considers the issue of operational efficiency and energy consumption, focusing on the difficulty of mechanical operation, both to find paths that make the energy efficient and easy to pass \cite{Chen6359075, Inotsume9006933}. Chen, et al. present a navigation method that consists of a formula for computing the directional-dependent cost related to slopes and a path planning algorithm for grid maps with slope cost \cite{Chen6359075}. Such a method can find a navigation path that improves the efficiency and adaptability for mobile robot in complex terrain environment. \cite{Hertle6698826} proposes a search-based path planning system for ground robots on terrain map. This system can effectively help ground robots travel through sloped ground. Based on hybrid A* search method, \cite{Brandao7511785} develops a footstep-planning algorithm for humanoids that is applicable to flat, slanted, and slippery terrain with high efficiency.

Besides, some scholars study obstacle avoidance path planning methods in dynamic slope environments, which are usually accompanied by real-time terrain recognition, 3D reconstruction of the map, and then planning feasible paths \cite{Kwok786612, Ugur9551617}. Kamaras, et al. present a path planning method to navigate a mobile robot to go through rugged, sloped, slippery, or, in general, challenging terrains \cite{Kamaras9288313}. The method produces smooth curves that are longer than a straight line to the goal, but are designed to be optimal (not longer than necessary) for a given maximum inclination that the robot platform is capable of. \cite{Linhui5360579} presents an obstacle avoidance path-planning algorithm based on stereo vision system for intelligent vehicle, such an algorithm uses stereo matching techniques to recognize the obstacles and planning path in real-time. \cite{Glaubit9483719} presents a neural network-based framework for the proactive planning and control of an autonomous mobile robot navigating through different terrains. Using such an approach, the mobile robot continually monitors the environment and the planned path ahead to accurately adjust its speed for successful navigation toward a desired goal. Obviously, these planning methods do not care about the global nature of the path.

In summary, the path planning methods considering slope mainly focus on the research of traversability and operability, some algorithms consider the energy consumption of the operation process, but planned paths are always local minimum. Thus, the planning methods considering the energy consumption and distance of the global path have not been widely studied. Previously,  Huang proposes a multi-objective planning method based on heuristic planning on 2.5D map (H3DM) \cite{huang20213}. In this method, a novel weight decay model is introduced to solve the modeling problem of heuristic function, then realizing the path planning with better trade-off of distance and energy consumption. However, the solutions obtained by this method is not good enough considering the global nature and the search efficiency. According to the above analysis, the multi-objective path planning method considering the global nature and time efficiency should be further researched.

To this end, we present the DMOP method using the DRL theory in this paper. The DMOP is mainly composed of a deep Q net (DQN) which can finish the planning task with higher efficiency in comparison with other methods.

\subsection{Problem Statement}

Though the DQN has the great potential to solve path planning problem, two technical problems should be solved here: (1) Convergence problem of DQN in high state space; (2) Multi-objective path planning problem.

\begin{itemize}
	
\item{Convergence problem of DQN in high state space. The 2.5D map has to be constructed on relatively high resolution, resulting in a large map size. That presents a challenge as the DQN struggles to converge due to the high state space. Even for a size of 20*20, the DQN may fail to converge sometimes, let alone higher resolution such as 50*50. } 

\item{Multi-objective path planning problem. The goal of the path planning task is not only to find a path from a given start point to a target but also to find a globally optimal path with a short length and low energy consumption.} 

\end{itemize}

%\begin{figure}[htp]
%	\centering
%	\includegraphics[width=0.9\columnwidth]{Problem_statement1.jpg}
%	\caption{Path planning problem on map with huge size}
%\end{figure}

%\begin{figure}[htp]
%	\centering
%	\includegraphics[width=0.63\columnwidth]{Problem_statement2.png}
%	\caption{Multi-objective path planning}
%\end{figure}

\subsection{Contributions}
The contribution of this work is listed as below:

\begin{itemize}

	\item{Novel topic: A deep reinforcement learning-based multi-objective 2.5D path planning method (DMOP) is proposed for GVs, realizing a good trade-off on distance and energy consumption on off-road terrain environment. }
	
	\item{Useful findings: 1) The DMOP plans paths with significant high efficiency in comparison with other methods. 2) The DMOP has powerful reasoning capability that enables it to perform arbitrary untrained planning tasks.}
	
	\item{Solved two technical problems: 1) The convergence problem of DQN in high state space is solved by using hybrid exploration strategy and reward shaping theory. 2) To solve the multi-objective 2.5D path planning problem using DRL, an exquisite reward function is designed. Such a reward function is modeled with the information of terrain, distance, and border.}

\end{itemize}

%\subsection{Organization of this paper}
%The remainder of this paper is arranged as: Section \uppercase\expandafter{\romannumeral2} is the preliminaries; Section \uppercase\expandafter{\romannumeral3} gives the design of DQN; Section \uppercase\expandafter{\romannumeral4} provides the experiments;  At last, the conclusion and prospect are given in Section \uppercase\expandafter{\romannumeral5}. In addition, the nomenclatures are shown in TABLE \uppercase\expandafter{\romannumeral1}.

\section{Preliminaries}

\subsection{Environment for path planning}
The planning task is regarded as an Atari-like game, and we have built an environment, namely "Mario Looks for Mushroom". The goal of this game is to control Mario to find path from an initial position to the target where the mushroom is located. In this game, the controller is given eight actions corresponding to eight directions (see Figure 1). Then this planning task is understood as: to make the agent learn to find the mushroom with relatively less energy consumption and travel time. In addition, 2.5D off-road maps are used in this paper. All of them are Digital Elevation Map (DEM) in which the Z-axis denotes the altitude, the XOY plane represents the horizontal plane. The DEM is assumed that each sampling (pixel, \((x,y,z)\)) can be reached in the planning process.

\begin{figure*}[htp]
\centering
\includegraphics[width=2\columnwidth]{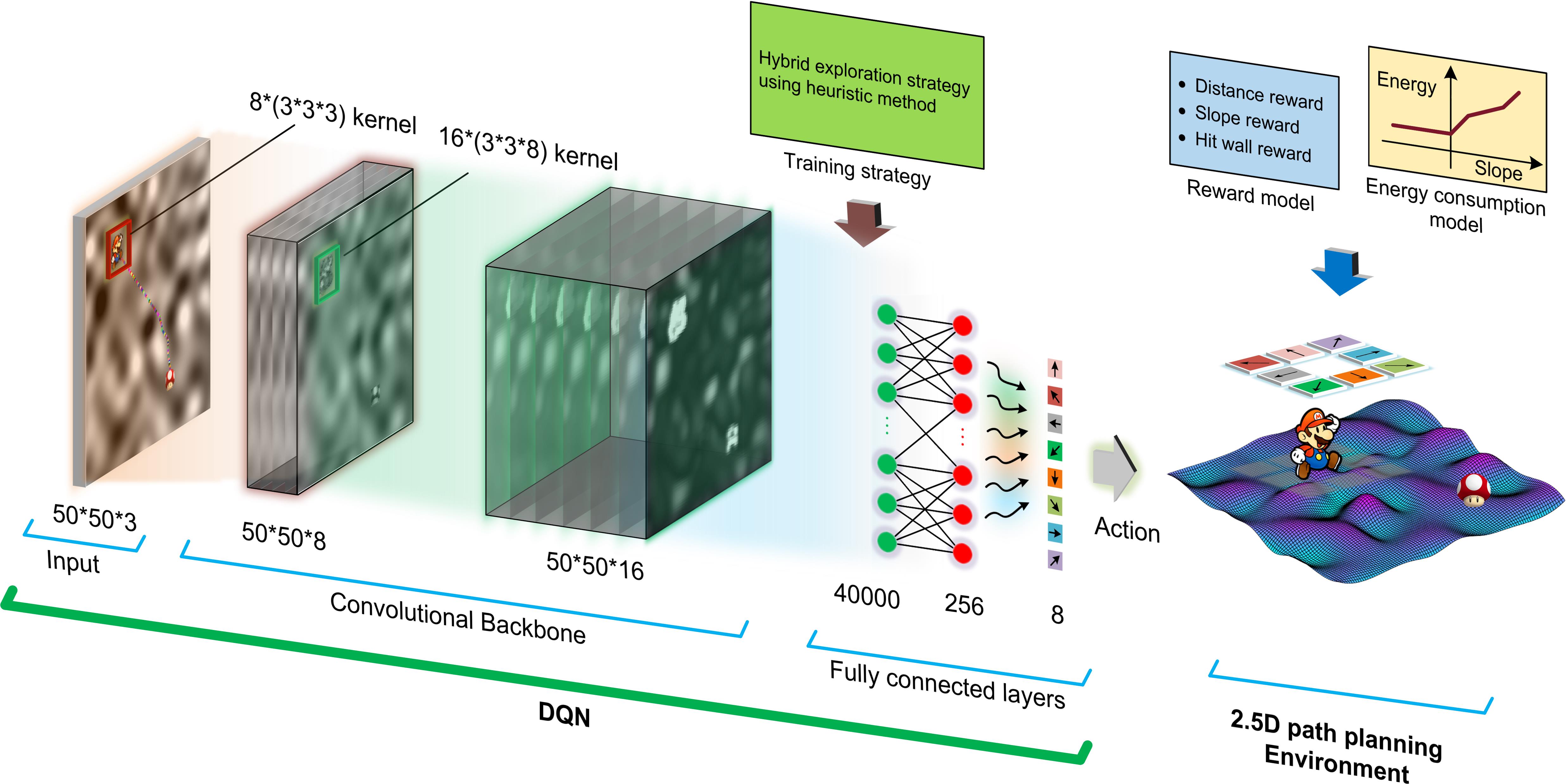}
\caption{DQN for 2.5D path planning task}
\end{figure*}

\newcommand{\tabincell}[2]{\begin{tabular}{@{}#1@{}}#2\end{tabular}}

\begin{table}[htp]
	\fontsize{7.5}{8}\selectfont   
	\centering  
	\renewcommand\arraystretch{1.8}
	\caption{Nomenclatures}
	\begin{tabular}{cl}
		\toprule[1.2pt] 
		Abbreviations & Meaning \\  
		\midrule
		\(P_n\) &  The \(n_{th}\) point in the planned path \\
		\cline{1-2}
		\(P_{n+1}\) &  The neighbor point of \(P_n\)\\
		\cline{1-2}
		\(P_d\) &  The destination (mushroom position) \\
		\cline{1-2}
		\(D(\cdot)\)   &  \tabincell{l}{Straight-line distance between two points} \\
		\cline{1-2}
		\(D_{total}\)   &  \tabincell{l}{Distance for a 2.5D path} \\
		\cline{1-2}
		\(E(\cdot)\) & \tabincell{l}{Energy consumption between two points }\\
		\cline{1-2}
		\(E_{total}\)   &  \tabincell{l}{Energy consumption for a 2.5D path} \\
		\cline{1-2}
		\(s\) &\tabincell{l}{ Observation of DQN (agent)}  \\
		\cline{1-2}
		\(a\) &\tabincell{l}{ Actions given by DQN (agent)}  \\
		\cline{1-2}
		\(a_h\) &\tabincell{l}{ Actions given by Heuristic method}  \\
		\cline{1-2}
		\(R_b\) &\tabincell{l}{Boundary penalty term in the reward function}  \\
		\cline{1-2}
		\(R_s\) &\tabincell{l}{Slope penalty term in the reward function}  \\
		\cline{1-2}
		\(R_t\) &\tabincell{l}{Reward term when Mario gets mushroom}  \\
		\cline{1-2}
		\(R_d\) &\tabincell{l}{Distance penalty term in the reward function}  \\

		\bottomrule[1.2pt]  
	\end{tabular}
\end{table}

\subsection{Distance calculation model for a path on 2.5D map}

In this paper, the distance \(D\) from point to point is calculated via the straight line distance equation, 
%补充三维坐标下两个点的直线距离公式。
\begin{equation}
\begin{split}
D(P_n,&P_{n+1})  = \\
&\sqrt{(x_n - x_{n+1})^2+(y_n - y_{n+1})^2+(z_n - z_{n+1})^2}
\end{split}
\end{equation}
where \( (x_n,y_n,z_n)\) is the 3D coordinate of \(P_n\), \( (x_{n+1},y_{n+1},z_{n+1})\) is the 3D coordinate of \(P_{n+1}\).

Hence, the distance of a specified path can be acquired as:
\begin{equation}
D_{total} = \sum_{i=1}^{d-1} D(P_i,P_{i+1})
\end{equation}
where \(d\) denotes the number of sampling point in a path.

\subsection{Energy consumption calculation for a path on 2.5D map}

Based on the above fitting curve and the distance calculation model, the energy consumption between two points is modeled as:
%给出两点之间的能量损耗计算公式
\begin{equation}
	\begin{split}
		E(P_n,P_{n+1}) = 
		\begin{cases}
			D(P_{n},P_{n+1}) \cdot E_{us}, z_{n} \leq z_{n+1}\\
			D(P_{n},P_{n+1}) \cdot E_{ds}, z_{n} \geq z_{n+1}
		\end{cases}
	\end{split}
\end{equation}
where \(E_{us}\) and \(E_{ds}\) are the up-slope and down-slope energy consumption estimation \cite{huang20213}. 

Then these two variables can be modeled as:
\begin{equation}
E_{us} = \beta / \frac{\pi}{4} \cdot \eta_{us}
\end{equation}
\begin{equation}
E_{ds} = 1 - \beta / \frac{\pi}{4} \cdot \eta_{ds}
\end{equation}
where \(\eta_{us}\) and \(\eta_{ds}\) denote the up-slope and down-slope estimated energy consumption per meter when \(\beta = 30^\circ\), respectively. \(\beta\) is calculated as:
\begin{equation}
\beta =\arctan(\frac{z_n-z_{n+1}}{\sqrt{(x_n - x_{n+1})^2+(y_n - y_{n+1})^2}})
\end{equation}
where \((x_n, y_n, z_n)\) and \((x_{n+1}, y_{n+1}, z_{n+1})\) are the co-ordinates of two reference point \(P_n\) and \(P_{n+1}\). In addition, the unit of \(E_{us}\) and \(E_{ds}\) is set as \(u\) in this paper. The unit of \(u\) is Joule, its value can be determined via test for a specific GV. Here, \(\eta_{us}\) and \(\eta_{ds}\) are set as 25 \(u/m\) and 0.25 \(u/m\), respectively.

Based on this model, the energy consumption of a path is calculated as:
\begin{equation}
E_{total} = \sum_{i=1}^{d-1} E(P_i,P_{i+1})
\end{equation}

\section{DQN-based multi-objective 2.5D path planning}
In this section, the design of the DQN, including the structure of the convolutional neural network (CNN), the modeling of the reward function and the training strategy as well as the path-enhanced method are introduced, respectively. The method built from these components ultimately solves the multi-objectives 2.5D path planning problem.

\subsection{Background of deep Q-learning}
The deep Q learning is a value-based deep reinforcement learning method which shows human-level control performance on playing Atari games \cite{mnih2015human}. From the perspective of reinforcement learning theory, the deep Q-learning evolves from Q-learning and deep learning. The Q-leaning always tries to maximize accumulated rewards through interacting with the environment by exploration. Generally, the current action of Q-learning contributes to future rewards to some degree. Thus, the future rewards are discounted using a factor \(\gamma\) for each time-step, and the future discounted return at \(t\) step is defined as:
\begin{equation}
R_t = \sum_{t=t'}^T \gamma^{t-t'}r_{t}
\end{equation}
where \(T\) is the final time-step for the control task. 

To get the policy that maximizes \(R_t\), the Q-leaning is apt to learning an optimal action-value function \(Q^*\) using a deep neural network:
\begin{equation}
Q^*(s,a)=\max_{\pi} \mathbb{E}\left[R_t |s_t=s, a_t=a, \pi\right]
\end{equation}
where \(\pi\) is a policy that maps a given sequential state to a sequential action. Because \(Q^*\) obeys to the Bellman equation, then \(Q^*\) can be written as:
\begin{equation}
Q^*(s,a)= \mathbb{E}_{s'\sim \varepsilon}\left[r+\gamma \max_{a'}Q^*(s',a')|s,a\right]
\end{equation}
where \(\varepsilon\) is the distribution of state \(s\). Based on this equation, \(Q^*\) can be estimated via iterative calculation: 
\begin{equation}
Q_{i+1}(s,a)= \mathbb{E}\left[r+\gamma \max_{a'}Q_i(s',a')|s,a\right]
\end{equation}
When \(i\rightarrow\infty\), \(Q_i\) will converge to \(Q^*\).

Here, a deep neural network is used to represent \(Q^*\), hence we call this type of Q-leaning as deep Q-leaning, and the deep neural network is called DQN. The loss function used to train the DQN is defined as:
\begin{equation}
L_i(\theta_i)=\mathbb{E}_{s,a\sim\rho(\cdot)}\left[\left(y_i-Q(s,a;\theta_i)\right)^2\right]
\end{equation}
where \(\theta_i\) is the parameter of network, \(\rho(\cdot)\) is the distribution of action \(a\), and \(y_i\) is defined as \(y_i= \mathbb{E}_{s'\sim \varepsilon}\left[r+\gamma \max_{a'}Q(s',a',\theta_{i-1})|s,a\right]\). This neural network is learned using the differentiation of \(L_i(\theta_i)\) with respective to \(\theta_i\):
\begin{equation}
\bigtriangledown_{\theta_i}L_i(\theta_i)=\mathbb{E}_{s,a\sim\rho(\cdot);s'\sim\varepsilon}\left[E_t \bigtriangledown_{\theta_i}Q(s,a;\theta_i)\right]
\end{equation}
where \(E_t = r+\gamma\max_{a'}Q_i(s',a';\theta_{i-1})-Q(s,a;\theta_i)\).

\subsection{Structure of DQN}
As shown in Fig.1, the adopted DQN is a convolutional neural network which is composed of three parts: input layer, convolutional backbone, fully connected layers. 

The input layer is the observation state of the DQN and is represented by a three-channel image (50*50*3). In which Mario, mushroom and the 2.5D map are drawn on the different channels, respectively. 

The convolutional backbone is built with 2 convolutional layers: The size of the convolutional kernel for the first layer is 8*(3*3*3), and the second is 16*(3*3*8). The activation function is ReLu. It should be mentioned that Mario and mushroom are used to replace the single pixel of current and target point, so that it makes the network easier to recognize the current location of agent and target, accelerating the learning process. In addition, no pooling layers are used here. According to numerous experiments, using the pooling layer in this structure can lead to a severe convergence problem. The reason may be that too much information is lost after pooling process.

The fully connected layers consist of 3 layers. The first layer connects to the output of the last convolutional layer. The second one has 256 nodes, both this layer and the former layer also use ReLu as the activation function. The last layer has 8 nodes that represent the Q value of actions. And it takes a linear function as the activation function.

\subsection{Design of reward function}

The reward function plays the key role on training the DQN. When the size of the map becomes bigger and bigger, the DQN will be more difficult to converge. If the agent only gets the reward at the target, the DQN will hardly converge through random exploration, let alone realizing the multi-objective planning. To solve this problem, an exquisite reward function is designed for the 2.5D multi-objective path planning task. Based on the reward shaping theory \cite{mnih2015human}, this function is built from the following points:
\begin{itemize}
	\item{The DQN aims to find a policy that maximizes the discounted accumulated reward (score), and the goal of this game is to find a path with a good trade-off between distance and energy consumption, hence the reward for Mario's step should be negative (penalty), otherwise, The DQN will tend to plan a path that maximizes the reward, even if it means making the path as long as possible. }
	\item{To guide Mario to find mushroom, a suitable distance-based reward should be provided. In addition, the agent can receive a bigger reward as it gets closer to the target. }
	\item{Considering the energy consumption in path planning, a slope-based penalty should be given.} 
	\item{To prevent the agent from cheating to get a higher score by hitting the wall, the hit-wall penalty should be relatively strong. }
	\item{A relatively big reward should be given to encourage agent to remember the experience when reaches the target. }
\end{itemize}

Based on the above ideas, the reward function is modeled as:
\begin{equation}
r = r_d + r_s + r_w + r_t
\end{equation}
where \(r_d\) is the distance penalty, \(r_s\) is the slope penalty, \(r_w\) denotes the hit-wall penalty and \(r_t\) stands for the reached reward. 

\subsubsection{Model of the distance penalty}
The model of distance penalty include two parts:
\begin{equation}
	r_{d} = r_{d1} + r_{d2}
\end{equation}
where \(r_{d}\) is a global distance penalty, \(r_{d1}\) is a step distance penalty and modeled as: 

\begin{equation}
	r_{d1} = - D(P_{n}, P_{n+1}) / D_{b1}
\end{equation}
where \(D(P_{n}, P_{n+1})\) is a step distance in 2.5D aspect, \(D_{b1}\) is a factor of the intensity, and \(D_{b1} = 10\) according to a bunch of tests.

Only \(r_{d1}\) is not enough as it neglects the terrain information, so the second part \(r_{d2}\) is introduced to compensate the penalty. \(r_{d2}\) heuristically penalizes the action that lead the next current position too far away from the destination in the 2D aspect. It is modeled as: 
\begin{equation}
	r_{d2} = - D_2(P_{n+1}, P_d) / D_{b2}
\end{equation}
where \(D_2(P_{n+1}, P_d)\) means the 2D distance between the next current position and the destination, \(D_{b2}\) is an intensive factor, its value is set according to the size of the map. Here, \(D_{b2} = 50\).

%The Gaussian function is used to model this term. For instance, the target is set at the center of the map (see Fig.6), the penalty quickly becomes smaller as Mario get close to the target, which encourages Mario to walk to the target. Here, \(r_d\) is calculated as:
%\begin{equation}
%r_d = e^{(-d^2/\eta)}-1;
%\end{equation}
%where \(d\) is the distance on x-y plate between the Mario and mushroom, \(\eta\) is a factor that controls the penal intensity of distance to the target.

%\begin{figure}[htp]
%\centering
%\includegraphics[width=0.85\columnwidth]{DistantPenalty.eps}
%\caption{Distant penalty}
%\end{figure}

\subsubsection{Model of the slope penalty}
The slope penalty can be calculated based on Eq.(3). Here, the slope penalty is defined as:
\begin{equation}
r_s = - E(P_{n-1}, P_{n}) \cdot k;
\end{equation}
where \(k\) is the intensity factor. \(k\) is set as 5 according to many training experiment. It is easy to understand that the penalty is proportional to energy consumption.

Here, a case is given to illustrate the slope penalty term. For a path depicted in Fig.2 a), the up-slope motion consumes lots of energy (see Fig.2 b)). Besides, we can control the intensity of slope penalty with different factor \(k\) (see Fig.2 c)). From Fig. 2 d), we can see that when \(k\) becomes bigger, the planned path are more sensitive to the slope and tries to avoid going up-slope.

\begin{figure}[htp]
	\centering
	\includegraphics[width=0.98\columnwidth]{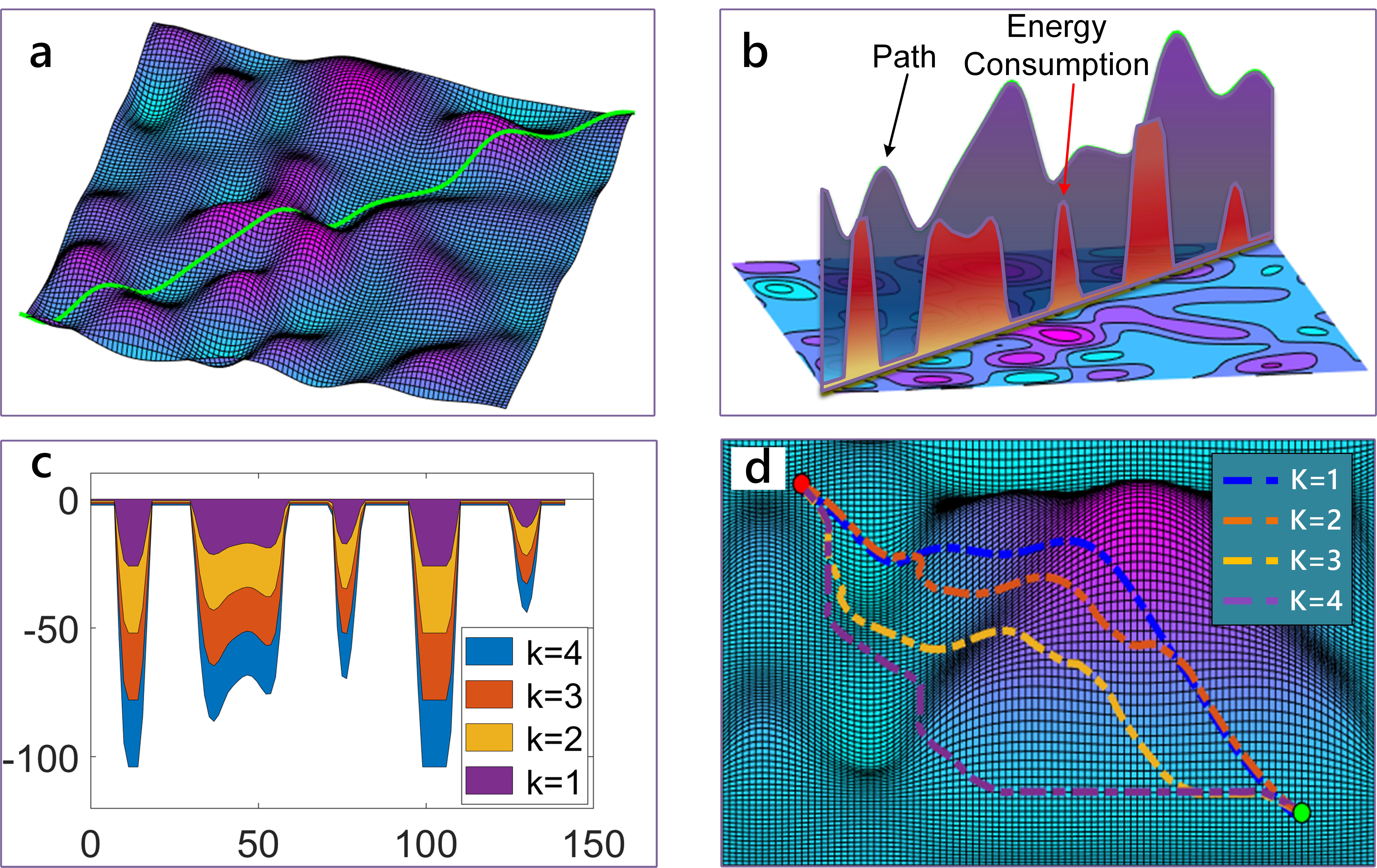}
	\caption{The function of slope penalty term. a) depicts a path for demonstration. b) shows the curves of energy consumption compared with the path. c) gives the energy consumption with different factor \(k\). d) is the corresponding planned results.}
\end{figure}

%\begin{figure}[htp]
%	\centering
%	\includegraphics[width=0.85\columnwidth]{SlopeEnergy2.jpg}
%	\caption{Corresponding energy consumption}
%\end{figure}
%
%\begin{figure}[htp]
%	\centering
%	\includegraphics[width=0.85\columnwidth]{SlopeEnergy3.eps}
%	\caption{Slope penalty with different intensities}
%\end{figure}

%\begin{figure}[htp]
%	\centering
%	\includegraphics[width=0.9\columnwidth]{Sharp_Factor.png}
%	\caption{Slope penalty with different intensities}
%\end{figure}

\subsubsection{Model of the hit-wall penalty}
The wall penalty term \(r_w\) is used to prevent the agent from hitting wall for higher reward, cheating in the game. \(r_w\) is modeled as:
\begin{equation}
r_w = \min((r_{wx} + r_{wy}), b) \cdot \gamma;
\end{equation}
where \(b\) is the upper bound, and \(b = 6\). \(r_{wx}\) and \(r_{wy}\) are corresponding penalty to the \(x\) and \(y\) axes, respectively. \(\gamma\) denotes the penalty intensity, and \(\gamma = 5\).  \(r_{wx}\) and \(r_{wy}\) are defined in the similar way:
\begin{equation}
\begin{split}
r_{wx} = 
	\begin{cases}
		(l_1-P_x) \cdot 2, P_x \leq l_1\\
		(P_x-l_2) \cdot 2, P_x \geq l_2\\
		\quad \quad \quad 0,\quad \quad others
	\end{cases}
\end{split}
\end{equation}
\begin{equation}
\begin{split}
r_{wy} = 
\begin{cases}
	(l_1-P_y) \cdot 2, P_y \leq l_1\\
	(P_y-l_2) \cdot 2, P_y \geq l_2\\
	\quad  \quad \quad 0,\quad \quad others
\end{cases}
\end{split}
\end{equation}
where \(l_1\) and \(l_2\) are the specific lines that start the wall penalty, and \(l_1 = 3, l_2 = 97\). \(P_x\) and \(P_y\) are the corresponding coordinates on the XOY plate.
%if x <= 10
%	bx = (10-x)*2;
%elseif x >= 90
%	bx = (x-90)*2;  
%else
%	bx = 0;   
%end
%\begin{figure}[htp]
%	\centering
%	\includegraphics[width=0.85\columnwidth]{WallPenalty.eps}
%	\caption{Wall penalty}
%\end{figure}
\subsubsection{Model of the reached reward}
The reached reward \(r_t\) is used to encourage the agent to remember the experiences it has encountered.
\begin{equation}
	r_t = 5 \cdot k
\end{equation}
where \(k\) is the same factor used in Eq. (18). The coefficient "5" is tuned manually according to the training results.

\subsection{Hybrid exploration strategy using heuristic method}
Since the state space is too big, the DQN is difficult to converge through random exploration, even a well-designed reward function is used. Here, hybrid exploration strategy is applied to improve the training performance. We borrow the idea of hybrid exploration strategy that uses the heuristic method \cite{ferguson2005guide} to generate useful experience for training the DQN. Thus, the DQN may select a recommended action given by the heuristic method in the training.

\begin{figure}[htp]
	\centering
	\includegraphics[width=0.65\columnwidth]{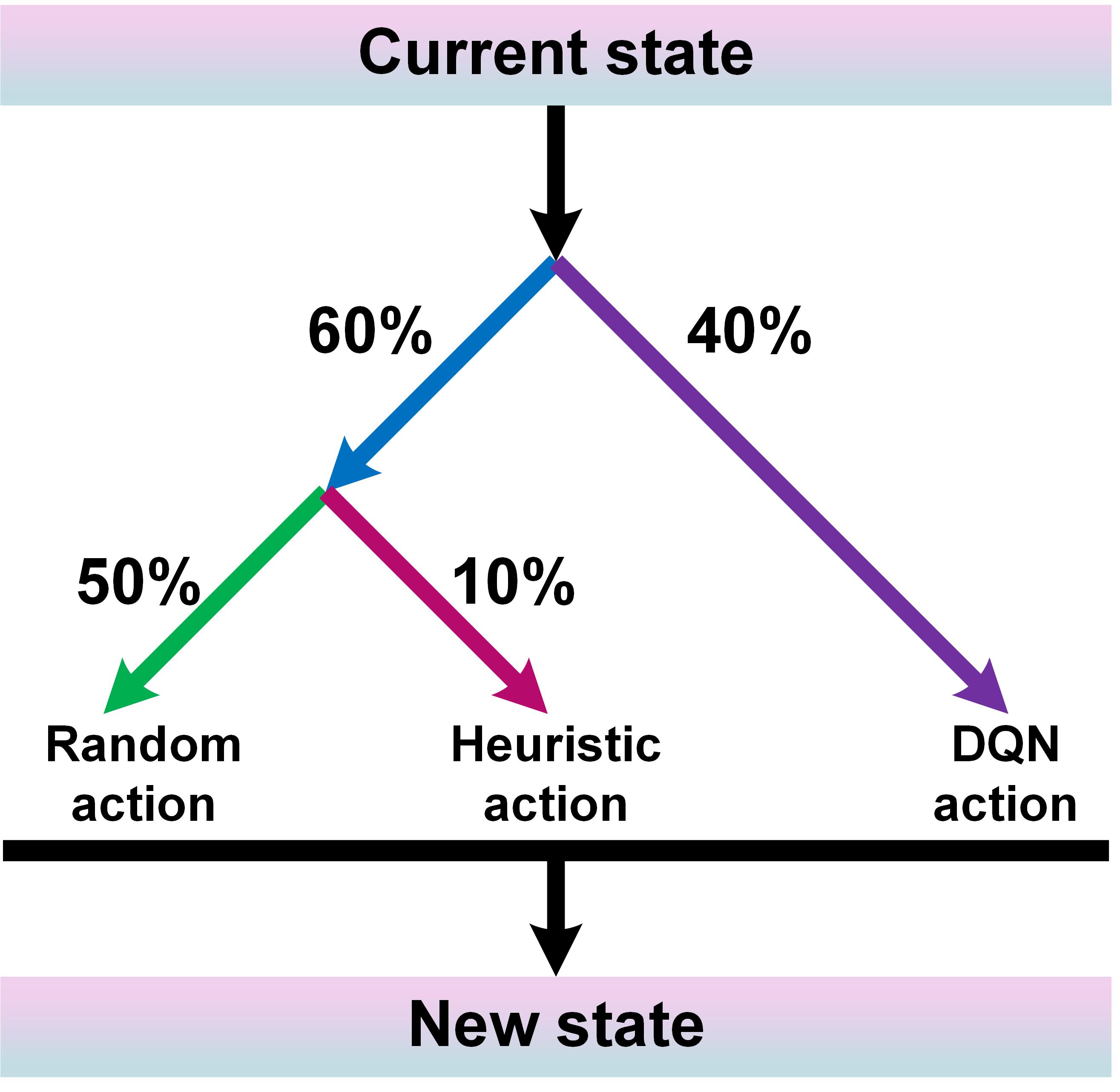}
	\caption{Action selecting in exploration}
\end{figure}

For the implementation of the heuristic method, we compare the distance between each neighbor point and the target, the action corresponding to the minimum value is selected as the recommended action. Though this action neglects the energy consumption, there is also a high probability of taking other actions (see Fig.3). As the training goes further, the exploration probability will decrease to a relatively small value (the number 60\% will go to 1\%, and 40\% will go to 99\% at the end). Eventually, the agent acquires the knowledge of how to plan paths.

%\begin{figure}[htp]
%	\centering
%	\includegraphics[width= 3.2 in]{A_star_algorithm.eps}
%	\caption{ Flow chart of A-star algorithm}
%\end{figure}
\subsection{Path-enhanced method}

The planned path on a lower resolution map includes many straight segments which should be curves on the higher resolution map, these segments can be recovered to curve. Here, a path-enhancing method is proposed to transform a lower resolution path (50*50) into a higher one (100*100, even higher).

\begin{figure}[htp]
	\centering
	\includegraphics[width=0.9\columnwidth]{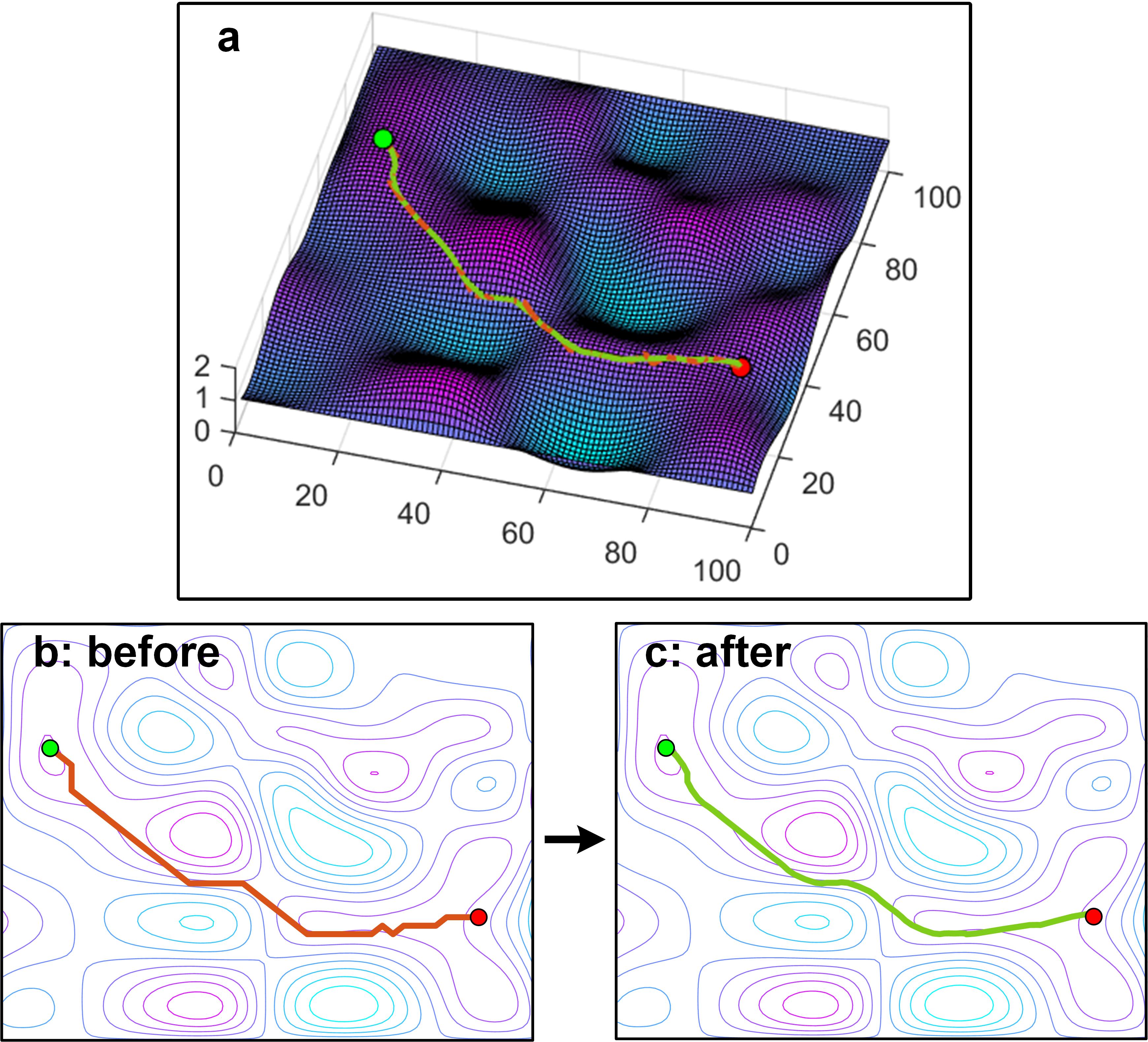}
	\caption{Example of path-enhanced. a) is the comparison on the 2.5D perspective. b) and c) are respectively the original path and the processed path on the 2D perspective.}
\end{figure}

The proposed method includes three steps:(1) Map the planned path into the corresponding high resolution map. (2) Extract the tuning points using the slope information of the path. (3) Connect the extracted points using the cubic spline interpolation method. Fig. 4 shows an example of a smooth process using the path-enhanced method. It shows that the green curve becomes smoother than the red curve. 

\subsection{Hyperparameters setting and Training result}
The proposed DMOP method is trained with multitask which means the start point and the target are set randomly in different episodes. The hyperparameters are listed in the following table.

\begin{table}[htp]
	\fontsize{7.5}{8}\selectfont   
	\centering  
	\renewcommand\arraystretch{1.8}
	\caption{Nomenclatures}
	\begin{tabular}{cc}
		\toprule[1.2pt] 
		Hyperparameters & Value \\  
		\midrule
		Learning rate & 0.0001 \\
		\cline{1-2}
		Discount factor & 0.99\\
		\cline{1-2}
		Batch size &  64 \\
		\cline{1-2}
		Batch number per training   &  20 \\
		\cline{1-2}
		Maximum episodes   &  6000 \\
		\cline{1-2}
		Maximum planning step per episode & 150\\
		\cline{1-2}
		Training interval   &  5 \\
		\cline{1-2}
		Initial probability of random action & 0.5 \\
		\cline{1-2}
		Initial probability of heuristic action  & 0.1 \\
		\cline{1-2}
		Initial probability of DQN action  & 0.4 \\
		\cline{1-2}
		Energy factor &  4 \\
		
		\bottomrule[1.2pt]  
	\end{tabular}
\end{table}

In Fig.5, the curves of reward in many cases are depicted. This result reveals that the convergence problem of DQN with high state space \cite{mnih2015human} can be solved via our method. When the DQN is converged, it learns how to recognize Mario and mushroom. In addition, we found an interesting phenomenon that the bottom of valleys (see the yellow circuits in Fig.5 b)) is regarded as the peak by DQN, which means the agent takes as much energy to cross a valley as it does to cross a mountain. 
 
\begin{figure}[htp]
	\centering
	\includegraphics[width=0.98\columnwidth]{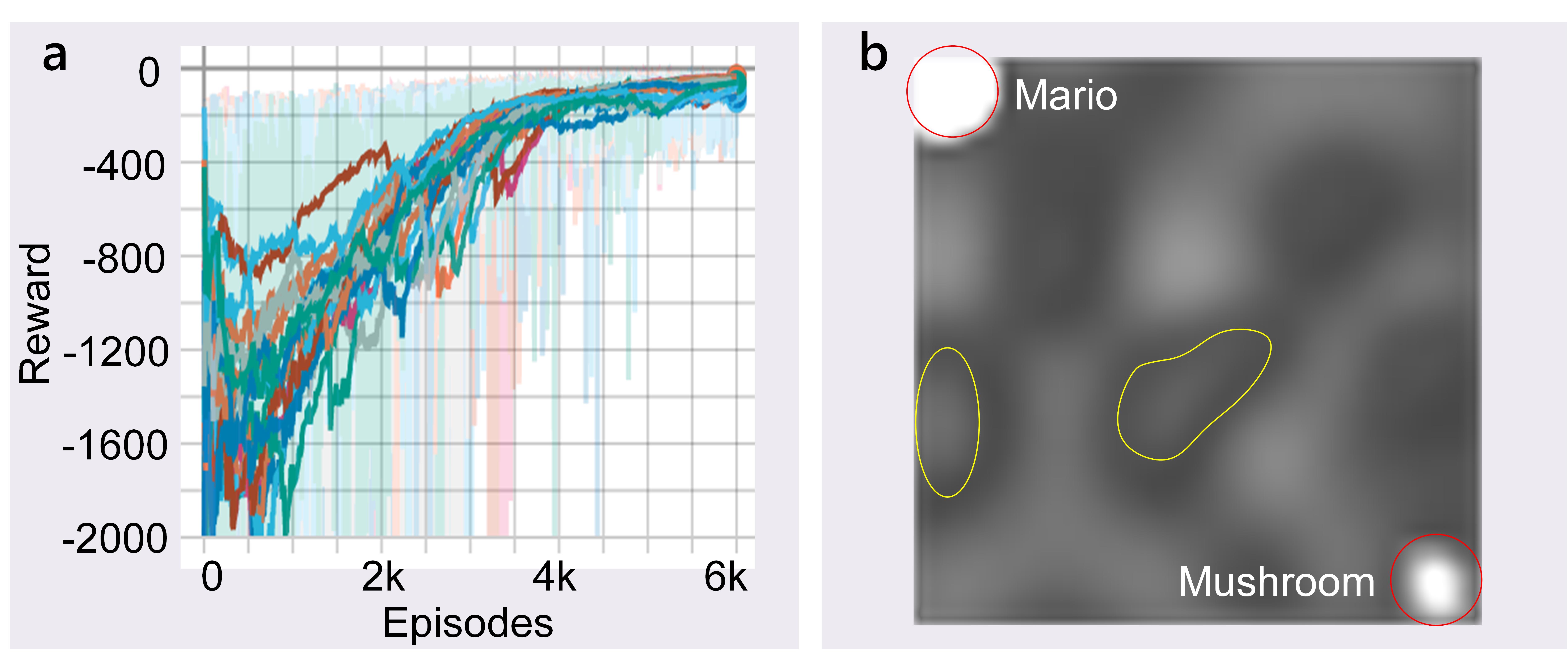}
	\caption{Result of training. a) the curves of reward. b) feature map of CNN }
\end{figure}

%\begin{figure}[htp]
%	\centering
%	\includegraphics[width=0.60\columnwidth]{Feature_map.png}
%	\caption{Feature map after training}
%\end{figure}

\section{Simulation and analysis}

\subsection{Setting and results}
In this section, 5 simulations have been carried out to verify our method. The 2.5D map used in this section is set as: both of its x and y axes range in [1, 100], and z-axis ranges in [0, 2]. This map can be regarded as it is zoomed from a 10 km * 10 km area with a maximum altitude of 200 m. Furthermore, the proposed method is compared with the H3DM method\cite{huang20213}, an improved RRT method \cite{Ugur9551617}, Dijkstra method \cite{Nie8991502}, and \({A^*}\) method \cite{Raja8023669}. All these methods are planning for multi-objective, where the energy consumption and path distance are using the same weights in comparison. All the methods run on a computer with NVIDIA GTX 3060 GPU, AMD Ryzen 5600 CPU, 16G RAM, and the codes are implemented with python. It should be mentioned that the DQN in DMOP method is trained with multitask, and all the tested tasks have not been used in the training process.

The results are shown in Fig.6 to Fig.10. Among these figures, a) and b) are the planned paths on 2.5D perspective, c) and d) are the corresponding result on 2D perspective, e), f) and g) illustrate the searching area of the Dijkstra method, A* method and RRT method, respectively. Meanwhile, the quantified results are shown in TABLE \uppercase\expandafter{\romannumeral3} to TABLE \uppercase\expandafter{\romannumeral7}.

	%------------
\begin{figure} 
	\centering
	\includegraphics[width=0.98\columnwidth]{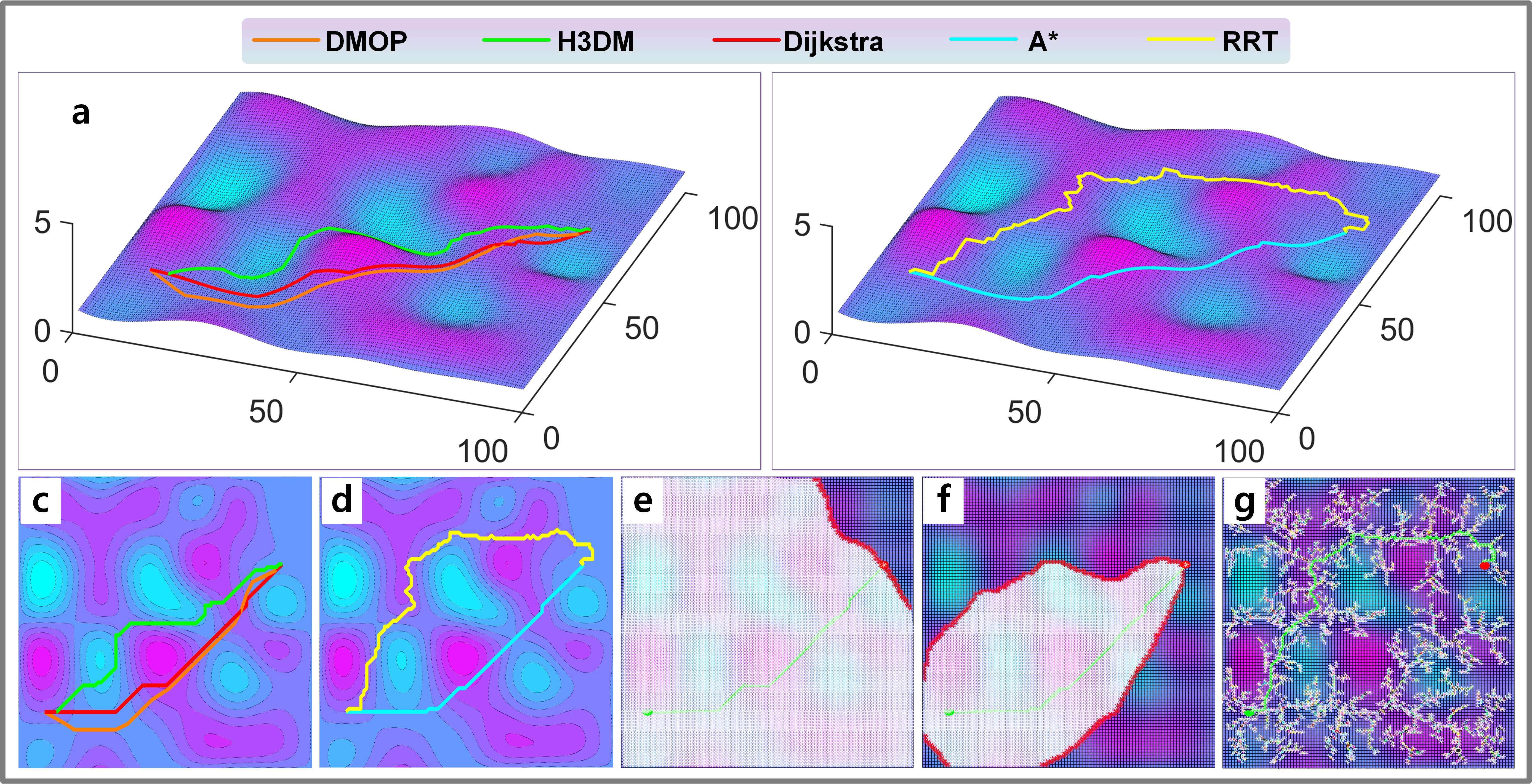}
	\caption{ Case 1 } 
	\label{fig} 
\end{figure}

%------------
\begin{figure}
	\centering
	\includegraphics[width=0.98\columnwidth]{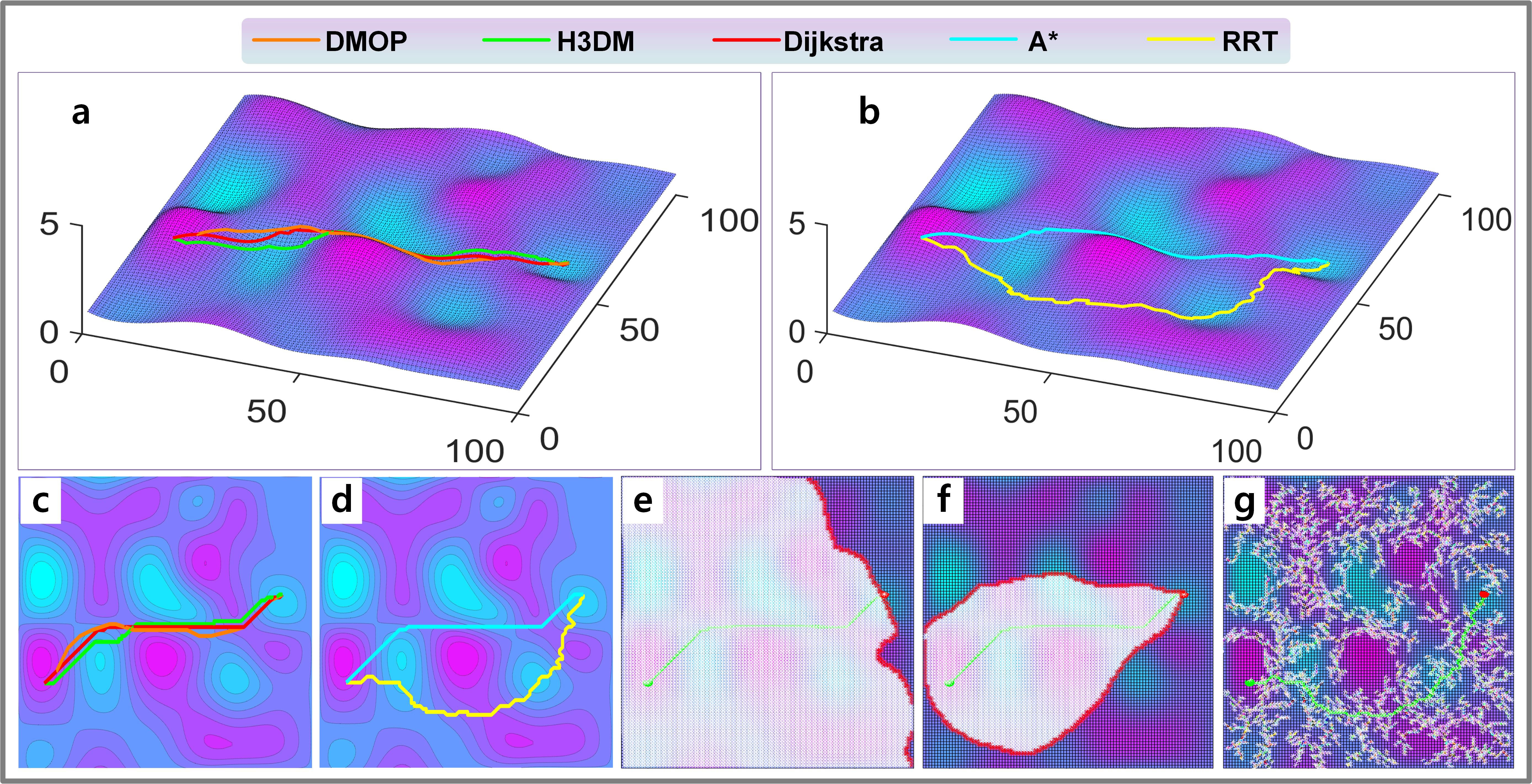}
	\caption{ Case 2 } 
	\label{fig} 
\end{figure}

%------------
\begin{figure}
	\centering
	\includegraphics[width=0.98\columnwidth]{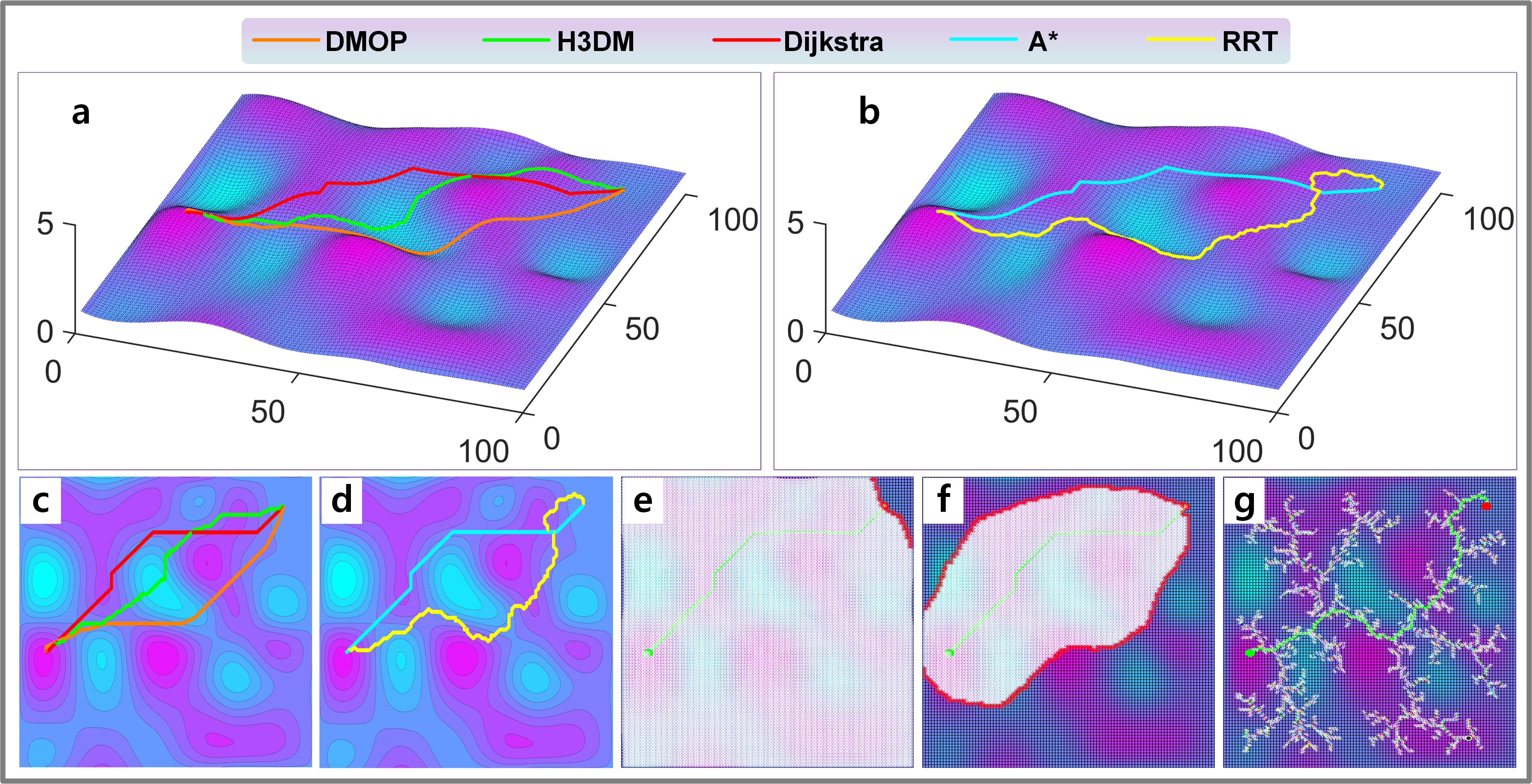}
	\caption{ Case 3 } 
	\label{fig} 
\end{figure}

%------------
\begin{figure} 
	\centering
	\includegraphics[width=0.98\columnwidth]{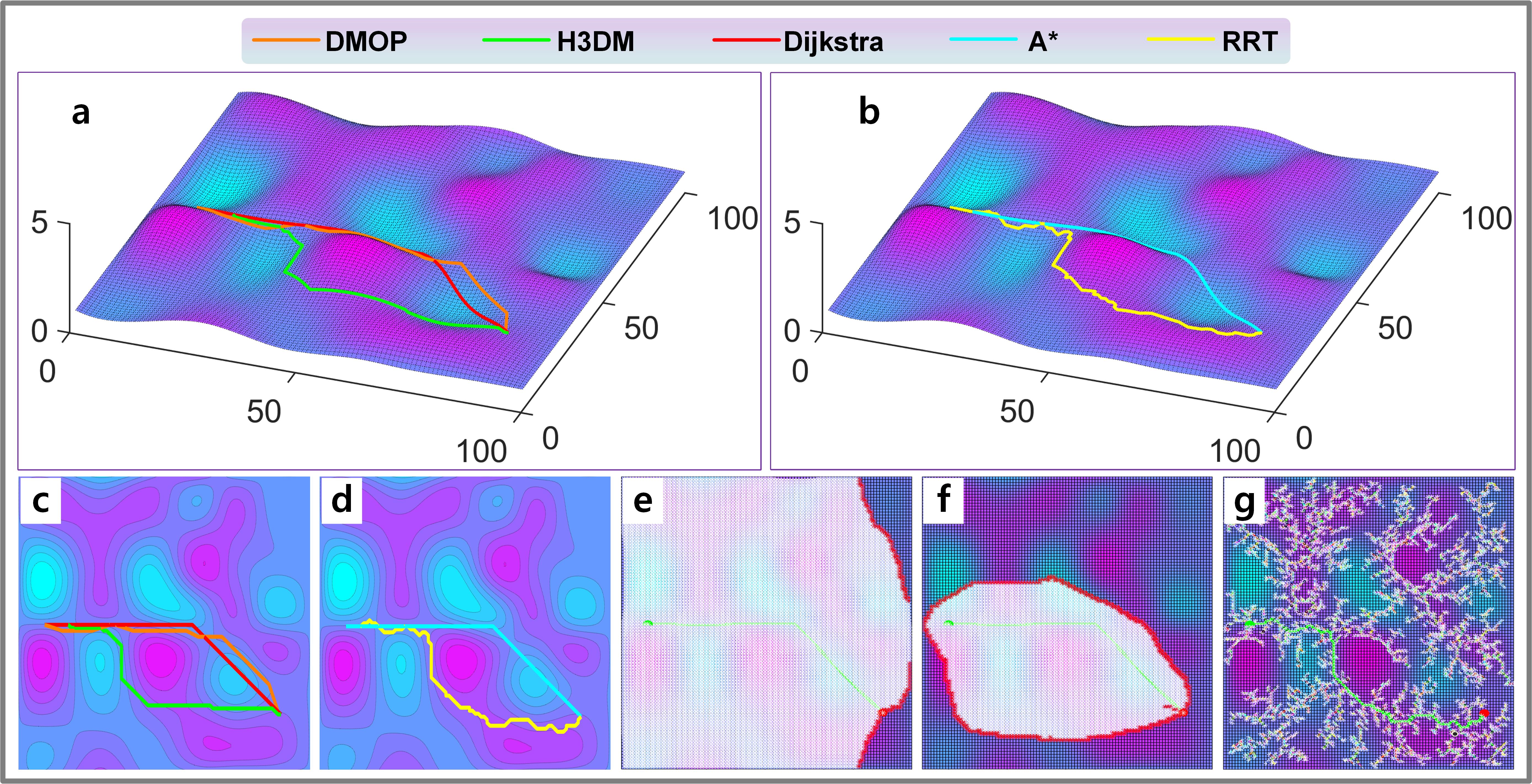}
	\caption{ Case 4 } 
	\label{fig} 
\end{figure}

%------------
\begin{figure} 
	\centering
	\includegraphics[width=0.98\columnwidth]{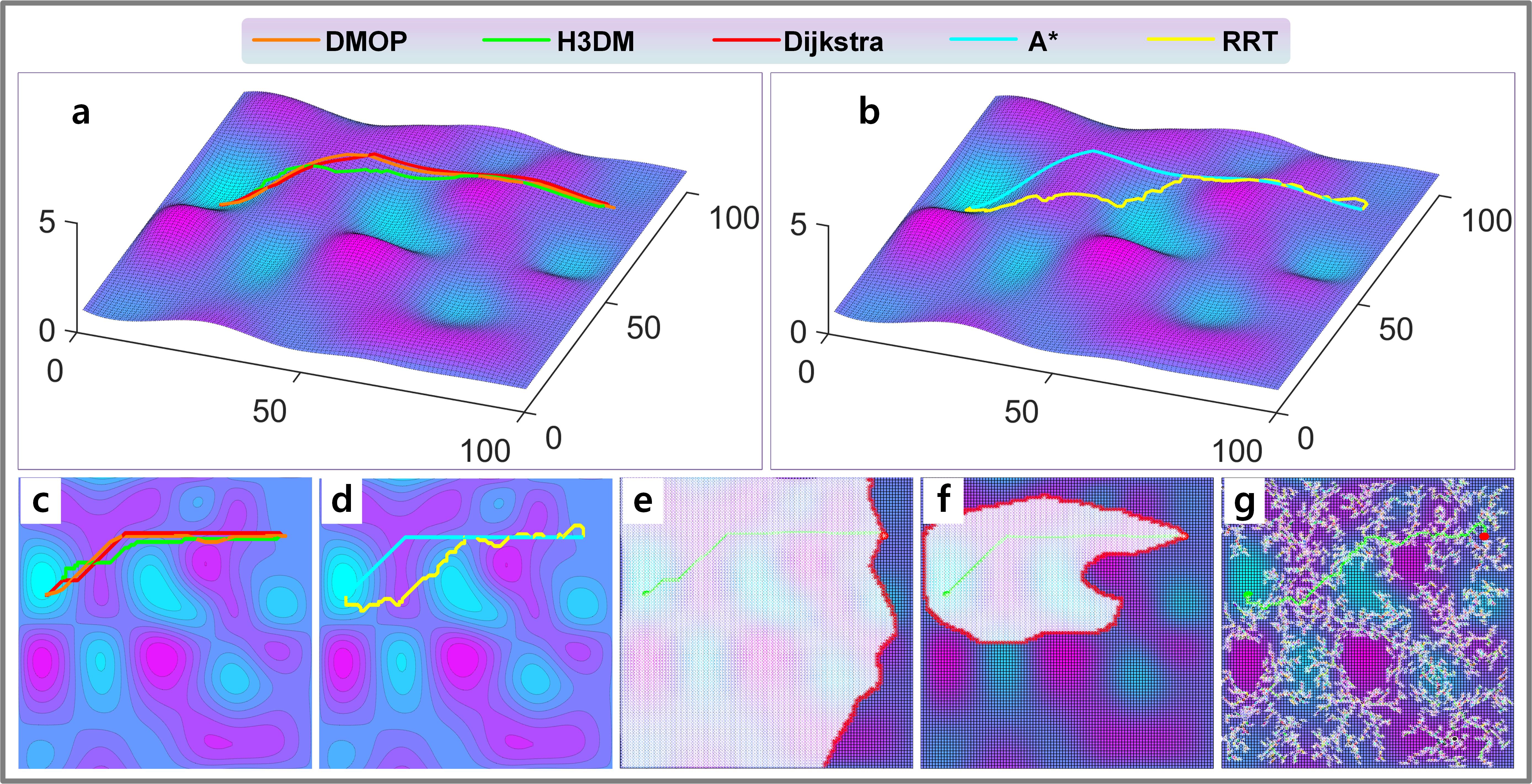}
	\caption{ Case 5 } 
	\label{fig} 
\end{figure}

\begin{table}[htp]
	\fontsize{8}{8}\selectfont   
	\centering  
	\renewcommand\arraystretch{1.5}
	\caption{Quantitative result of Case 1}
	\begin{tabular}{lcccc}
		\toprule[1.2pt] 
		\textbf{Method}  & \textbf{Distance (m)} & \textbf{Energy (u)} 	& \textbf{Sum} & \textbf{Time (s)}  \\
		\midrule
%		\multirow{6}{*}{Case 1}
		DMOP  & 10686.96  & 13027.12 & 23714.08  & \textbf{0.32} \\
		H3DM  & 10956.94  & 13832.42 & 24789.36  & 10.49          \\
		Dijkstra  & 10195.35  & 13073.7 & 23269.05  & 133.63    \\
		\(A^*\) & 10195.08 & 13076.01 & 23271.09  & 53.28   \\
		RRT  & 15337.99 &	18897.25 &	34235.25 &	2415.35     \\
		\bottomrule[1.2pt]
	\end{tabular}
\end{table}

\begin{table}[htp]
	\fontsize{8}{8}\selectfont   
	\centering  
	\renewcommand\arraystretch{1.5}
	\caption{Quantitative result of Case 2}
	\begin{tabular}{lcccc}
		\toprule[1.2pt] 
		\textbf{Method}  & \textbf{Distance (m)} & \textbf{Energy (u)} 	& \textbf{Sum} & \textbf{Time (s)}  \\
		\midrule
		%		\multirow{6}{*}{Case 1}
		DMOP  & 9453.30  & 11197.33 & 20650.63  & \textbf{0.33}        \\
		H3DM  & 9358.24  & 11884.20 & 21242.44  & 11.92        \\ 
		Dijkstra  & 9251.96	& 11357.54 & 20609.50 & 100.83     \\
		\(A^*\) & 9251.81 & 	11393.83& 	20645.64& 	38.38  \\
		RRT  & 12388.25 &	16331.19 &	28719.44 &	3271.31   \\
		
		\bottomrule[1.2pt]
	\end{tabular}
\end{table}

\begin{table}[htp]
	\fontsize{8}{8}\selectfont   
	\centering  
	\renewcommand\arraystretch{1.5}
	\caption{Quantitative result of Case 3}
	\begin{tabular}{lcccc}
		\toprule[1.2pt] 
		\textbf{Method}  & \textbf{Distance (m)} & \textbf{Energy (u)} 	& \textbf{Sum} & \textbf{Time (s)}  \\
		\midrule
		%		\multirow{6}{*}{Case 1}
		DMOP  & 10355.95  & 12398.70 & 22754.65  & \textbf{0.32}     \\
		H3DM  & 10786.18  & 14104.28 & 24890.46  & 9.77        \\
		Dijkstra  & 10371.01& 12308.39	& 22679.40	& 100.97  \\
		\(A^*\)  & 10371.01  & 12308.39  &	22679.40  &	53.48   \\
		RRT  & 14490.30	& 19435.67 & 33925.97 & 958.51    \\
		
		\bottomrule[1.2pt]
	\end{tabular}
\end{table}

\begin{table}[htp]
	\fontsize{8}{8}\selectfont   
	\centering  
	\renewcommand\arraystretch{1.5}
	\caption{Quantitative result of Case 4}
	\begin{tabular}{lcccc}
		\toprule[1.2pt] 
		\textbf{Method}  & \textbf{Distance (m)} & \textbf{Energy (u)} 	& \textbf{Sum} & \textbf{Time (s)}  \\
		\midrule
		%		\multirow{6}{*}{Case 1}
		DMOP  & 9500.19  & 11754.78 & 21254.97  & \textbf {0.34}     \\
		H3DM  & 9890.28  & 12049.37 & 21939.66  & 12.97     \\
		Dijkstra  & 9246.16 & 	11444.45 & 	20690.61	& 100.97  \\
		\(A^*\)  & 9246.16 & 	11444.45 & 	20690.61 & 	40.37   \\
		RRT  & 11378.68	  & 14074.97  &	25453.65  &	2614.11  \\
		
		\bottomrule[1.2pt]
	\end{tabular}
\end{table}

\begin{table}[htp]
	\fontsize{8}{8}\selectfont   
	\centering  
	\renewcommand\arraystretch{1.5}
	\caption{Quantitative result of Case 5}
	\begin{tabular}{lcccc}
		\toprule[1.2pt] 
		\textbf{Method}  & \textbf{Distance (m)} & \textbf{Energy (u)} 	& \textbf{Sum} & \textbf{Time (s)}  \\
		\midrule
		%		\multirow{6}{*}{Case 1}
		DMOP  & 9017.33  & 12892.16 & 21909.49 & \textbf{0.29}     \\
		H3DM  & 9116.10  & 13127.90 & 22244.23  & 10.48        \\
		Dijkstra  & 8917.89 &	12470.11 &	21387.99 & 97.07   \\
		\(A^*\)  & 8835.11 & 	12629.49 & 	21464.60 & 	39.91  \\
		RRT  & 11244.09	& 16680.88 & 27924.98 & 2784.60  \\
		
		\bottomrule[1.2pt]
	\end{tabular}
\end{table}

From the results, we can see that the planning speed of DMOP is super fast, which is around 0.32 seconds, and the quality of its results is very close to those of the Dijkstra method. The Dijkstra method can get the global optimal paths but needs about 100 seconds or more, thus its efficiency is low. The A* method often takes dozens of seconds to complete the planning. The RRT is very slow and unstable (Case 2 took 3271 seconds to find the target, whereas Case 3 took 958 seconds), the global nature of the planned results is also average, the path smoothness is poor. The time required for H3DM to complete the planning task is around 10 seconds, the global nature of the paths it plans is a bit worse than that of DMOP. Though DMOP does not find the global optimal solution, its path quality is close to that of A* method. Considering the search speed and the path quality, the efficiency of DMOP is the best among all the methods. 

Overall, if we assume that the paths planned by the DMOP, A*, and H3DM are all acceptable, then the DMOP is over 100 times faster than the A* method and about 30 times faster than the H3DM.

\subsection{Discussion}

The simulation results show that although Dijkstra method can find the global optimal path, it always searches almost the whole map to get the result, and the required time increases exponentially with the increase in map size. The A* method is recognized as the most efficient method for one objective path planning task on 2D maps, which takes less time and has a high quality of solution because of the added heuristic function. In the 2.5D multi-objective path planning task, the A* method still maintains a high efficiency. However, we made a special design of the heuristic function of the A* method according to the planning task of this paper to achieve relatively stable results. The special design is that we model its heuristic function with linear distance, curvilinear distance and energy consumption estimation, and set appropriate parameters. For the RRT algorithm, there is blindness and randomness in its search, which leads to its inefficiency and poor smoothness of the planned paths. From the figures of the results, we can see that the points searched by RRT are not as many as those of Dijkstra method, but its search area basically covers the whole map.

Whether it is the Dijkstra method, A* or RRT, they all need to search a certain area, causing the method to take much time. The H3DM is built based on the idea of pure heuristics, the algorithm tends to be greedy search, with the least number of search points for path planning, so its efficiency is much higher than the previous methods, and the method takes only about 10 seconds. Due to the extreme compression of the search space, it sacrifices the quality of the solutions to some extent. However, in general, the solution obtained from H3DM does not undergo significant sacrifices, and the time required to plan the path is further reduced, thus indicating an improvement.

Compared with these previous methods, the DMOP method has a relatively high quality of solution and a very short search time. For the global nature of the solution of DMOP, during the training process as DQN learns to judge the peaks and valleys, and it always sees the global view(the whole map as input), this makes it achieve the global path planning from another perspective. Furthermore, through many training tasks, DQN learns to assess the cost of each point in the planning process from a global perspective, and thus plan paths that are better globally. For the search time, theoretically, the DMOP method searches the same region as the H3DM method does, and both methods perform greedy selection planning from neighbor points each time. However, due to the complexity of the heuristic function model in the H3DM, it takes much time to estimate the cost of each neighbor point. On the other hand, the DMOP requires only simple forward operations, and the calculations inside the network  are straightforward, which makes the method very fast.

Furthermore, because DMOP is tested on the tasks that it has never seen before, another important conclusion we can draw from the simulation is that the method has powerful reasoning capability that enables it to perform arbitrary untrained planning tasks on the given map. On this basis, for exploration robots working on a specific terrain area, the DMOP method can learn to plan paths on this area in advance, and quickly generate the navigation path for these robots while they are working.

\section{Conclusion and prospect}
In this paper, a deep reinforcement learning-based multi-objective 2.5D path planning method is investigated for energy-saving of ground vehicles. Such a method can efficiently plan paths which has a good trade-off on distance and energy consumption. In the implementation, we solved two technical problems: Convergence problem of DQN in high state space, and realizing multi-objective path planning. For the first problem, we linearly transform the original map to a relatively lower resolution form without losing its terrain information, which alleviates the burden of the exploration of DQN. When DQN finishes planning, the planned path will be transformed to the original map using a path-enhancing method. Meanwhile, the hybrid exploration strategy is also used to speed up the training of DQN. For the second problem, an exquisite reward function is designed to guide the DQN to plan paths multi-objectively. The reward function is modeled with the information of terrain, distance, and border using the reward shaping theory. Simulation shows the proposed method is over 100 times faster than the A* and about 30 times faster than the H3DM method. Also, simulation proves the method has powerful reasoning capability that enables it to perform arbitrary untrained planning tasks on the map where the ground vehicles are required to operate. 

The future work will be carried out from 2 aspects: (1) to study a new deep reinforcement learning method that converges more quickly than the DQN. (2) to explore the application of deep reinforcement learning in solving other path planning problems.

%\section*{Acknowledgment}

%\ifCLASSOPTIONcaptionsoff
%  \newpage
%\fi



%\bibliographystyle{IEEEtran}

%\begin{thebibliography}{1}
%\end{thebibliography}

\bibliographystyle{IEEEtran}
\bibliography{IEEEabrv,IEEEexample}

\end{document}